\documentclass[letterpaper, 10 pt, conference]{ieeeconf}  

\IEEEoverridecommandlockouts                              

\usepackage{graphics} 
\usepackage{epsfig} 
\usepackage{draftwatermark}
\usepackage{amsmath} 
\usepackage{amssymb}  
\usepackage[ruled,linesnumbered]{algorithm2e}

\usepackage{color} 
\usepackage{multicol}
\usepackage{multirow}
\usepackage[normalem]{ulem}

\usepackage{tikz}
\usepackage{textcomp}
\usepackage{lipsum}

\newcommand\copyrighttext{%
  \footnotesize \textcopyright 2012 IEEE. Personal use of this material is permitted.
  Permission from IEEE must be obtained for all other uses, in any current or future
  media, including reprinting/republishing this material for advertising or promotional
  purposes, creating new collective works, for resale or redistribution to servers or
  lists, or reuse of any copyrighted component of this work in other works.}
 \newcommand\copyrightnotice{%
 \begin{tikzpicture}[remember picture,overlay]
 \node[anchor=south,yshift=10pt] at (current page.south) {\fbox{\parbox{\dimexpr\textwidth-\fboxsep-\fboxrule\relax}{\copyrighttext}}};
 \end{tikzpicture}%
 }

\title{\LARGE \bf
FBG-Based Control of a Continuum Manipulator \\ Interacting With Obstacles*
}

\author{Shahriar Sefati$^{1, 2, 3}$, {\it Student Member, IEEE}, Ryan J. Murphy$^{4}$, {\it Member, IEEE}, Farshid Alambeigi$^{1, 2}$, \\ {\it Student Member, IEEE}, Michael Pozin$^{2}$, Iulian Iordachita$^{1,2}$, {\it Senior Member, IEEE}, \\ Russell H. Taylor$^{1,3}$, {\it Life Fellow, IEEE}, and Mehran Armand$^{1,2,4}$, {\it Member, IEEE}
\thanks{$^{1}$S. Sefati, F. Alambeigi, I. Iordachita, R. H. Taylor, and M. Armand are with the Laboratory for Computational Sensing and Robotics, Johns Hopkins University, Baltimore, MD, USA \{sefati, falambe1, iordachita, rht\}@jhu.edu, mehran.armand@jhuapl.edu.
}%
\thanks{$^{2}$S. Sefati, F. Alambeigi, M. Pozin, I. Iordachita, and M. Armand are with the Department of Mechanical Engineering, Johns Hopkins University, Baltimore, MD, USA \{sefati, falambe1, mpozin1, iordachita\}@jhu.edu, mehran.armand@jhuapl.edu.
}%
\thanks{$^{3}$S. Sefati, and R. H. Taylor are with the Department of Computer Science, Johns Hopkins University, Baltimore, MD, USA \{sefati, rht\}@jhu.edu.
}%
\thanks{$^{4}$R. J. Murphy, and M. Armand are with the Johns Hopkins University Applied Physics Laboratory, Laurel, MD, USA (ryan.murphy@jhuapl.edu, mehran.armand@jhuapl.edu). R. J. Murphy is currently with Auris Health, Inc., Redwood City, CA.
}%
}

\begin{document}

\SetWatermarkText{Accepted for IROS 2018}
\SetWatermarkScale{0.4}

\maketitle
\vspace*{-1cm}
\thispagestyle{empty}
\pagestyle{empty}
\copyrightnotice

\begin{abstract}

Tracking and controlling the shape of continuum dexterous manipulators (CDM) in constraint environments is a challenging task. The imposed constraints and interaction with unknown obstacles may conform the CDM's shape and therefore demands for shape sensing methods which do not rely on direct line of sight. To address these issues, we integrate a novel Fiber Bragg Grating (FBG) shape sensing unit into a CDM, reconstruct the shape in real-time, and develop an optimization-based control algorithm using FBG tip position feedback. The CDM is designed for less-invasive treatment of osteolysis (bone degradation). To evaluate the performance of the feedback control algorithm when the CDM interacts with obstacles, we perform a set of experiments similar to the real scenario of the CDM interaction with soft and hard lesions during the treatment of osteolysis. In addition, we propose methods for identification of the CDM collisions with soft or hard obstacles using the jacobian information. Results demonstrate successful control of the CDM tip based on the FBG feedback and indicate repeatability and robustness of the proposed method when interacting with unknown obstacles.

\end{abstract}


\section{INTRODUCTION}
In recent years, continuum manipulators have been used for a variety of applications \cite{walker2016snake}-\cite{camarillo2004robotic}. Many of these applications require the manipulator to maneuver through a constrained environment which takes advantage of the flexibility of these manipulators. With limited prior knowledge of the constrained environment geometry and possible interactions with unknown obstacles, it is vital to choose a suitable shape sensing technique and control algorithm for controlling the CDM. A range of methods of shape sensing, modeling, and control strategies have been proposed depending on the geometry and curvature behavior of these manipulators \cite{webster2010design}-\cite{alambeigi2018acc}.  

\begin{figure}[t]
      \centering
      \includegraphics[scale=0.42]{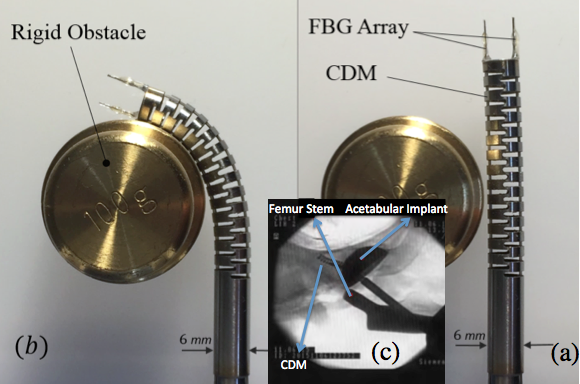}
      \caption{ The CDM in interaction with hard obstacle: a) straight configuration b) bending configuration c) x-ray image of the CDM inserted into the screw holes of the acetabular component}
      \label{fig:hard}
   \end{figure}

The constrained environments exist often in medical applications where the continuum manipulator may interact with tissues and organs. One such application is the robot-assisted treatment of osteolysis (bone degradation) occurring due to the wear of the polyethylene liner of the acetabular component after total hip replacement. For the less invasive treatment of this medical problem, we have previously developed a planar non-constant curvature cable-driven continuum manipulator (CDM) with a relatively large ($4$ mm) instrument channel (Fig. \ref{fig:hard}) \cite{murphy2014design}.

The CDM enables surgeons to access the region behind the well-fixed implant by passing the manipulator through the screw holes of the acetabular component \cite{murphy2014design,alambeigi2018acc}. A flexible debriding tool is then passed through the instrument channel of the manipulator to remove and clean the osteolytic lesion \cite{alambeigi2016design,alambeigi2017curved}. These application specifics demand the manipulator to be capable of reaching and maneuvering into the environment behind the implant. When the CDM is passed through the acetabular implant inside the patient body, it will not be visually accessible to the optical tracking systems or cameras, demanding a shape sensing technique that does not require direct line of sight. In addition, the CDM may interact with soft and hard tissues with unknown geometries and physical properties behind the acetabular implant, which makes the control of the CDM more challenging. 

Several approaches exist for shape and tip sensing of continuum manipulators. Examples of such sensing approaches include electromagnetic tracking, infrared optical tracking and cameras, image-based methods such as fluoroscopy and magnetic resonance, and fiber optic shape sensing methods \cite{sensingTechniques}-\cite{sefati2018ismr}. Each of these methods suffer from limitations such as magnetic interference, backlash, and the need for direct line of sight. In addition, there is limitation in extensive intraoperative use of image-based methods for real-time shape sensing due to radiation exposure. Fiber Bragg Gratings (FBG), however, provide real-time feedback without requiring a direct line of sight. Their small size, flexibility, and minimal effects on stiffness are advantages that make these sensors a great candidate for integration in continuum manipulators \cite{sefati2016fbg}, \cite{sefati2017fbg}.

With the CDM's shape and tip position provided by FBG shape sensors in real-time, an appropriate control strategy should be used for feedback control of the CDM. Previously, several kinematic and mechanical models have been developed for various types of continuum manipulators \cite{camarillo2009task}-\cite{mahoneyreview}. These models, however, cannot account for interactions with the environment and the possible effects on the control and kinematics of the manipulator. Considering the interactions of the CDM with a constrained unknown environment in the case of osteolysis, an optimization-based control strategy incorporating the physical constraints on the CDM and FBG positional information can potentially be a practical approach towards controlling the CDM \cite{yip2014model}.

This paper develops an optimization-based control algorithm with optimization constraints based on the osteolysis application requirements and CDM physical constraints, independent of the mechanical model of the CDM and prior knowledge of the workspace environment. Considering this algorithm, using the measurements of the actuators (cables) and constructing the shape and tip position of the CDM in real-time by the FBG shape sensors, we update the manipulator Jacobian. This Jacobian is then used to find control input commands to manipulate the CDM to a desired target point. We evaluate performance of the proposed method on the CDM both in free and constrained environments containing unknown obstacles with different properties. In addition, we thoroughly investigate the behavior of the updated jacobian when the CDM interacts with objects, and identify CDM collisions with hard or soft objects using this information. 

The contributions of this paper are threefold: 1) detailed CDM shape reconstruction using a novel large deflection FBG-based shape sensor that consists of a three open-lumen Polycarbonate tube as substrate, 2) development of an optimization-based control strategy using FBG tip position feedback for controlling the CDM in environments with obstacles, 3) identification of CDM collisions with soft and hard obstacles using the updated jacobian information.  

\section{METHODS} 

\subsection{CDM Specifications}

The CDM developed for the osteolysis application is constructed of two nested pieces of nitinol tubing with an outer diameter of 6 mm, designed to fit through the screw holes of an acetabular implant (Fig. \ref{fig:CDM}) \cite{murphy2014design}. Nitinol is chosen due to its super-elastic property which grants the CDM more flexibility. In addition, the CDM consists of a $4$ mm  instrument channel for passing tools such as debridering tools, endoscope and suctioning mechanisms \cite{alambeigi2016design}. The CDM wall consists of channels for passing the actuation cables and the shape sensing units (described in section \ref{Feedback}). One shape sensing unit is inserted through the channel on each side of the CDM for accurate CDM tip estimation (Fig. \ref{fig:CDM}).

\subsection{Shape Sensing Unit} \label{Feedback}

\begin{figure}[bt]
      \centering
      \includegraphics[scale=0.51]{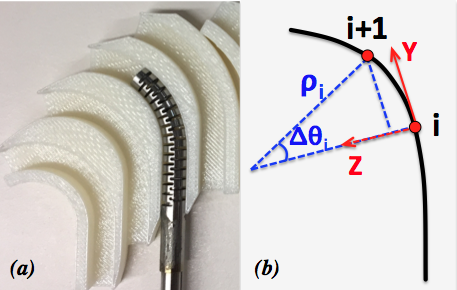}
      \caption{a) CDM curvature calibration jig and b) step-wise shape reconstruction for a small segment $\Delta s$ (between exaggerated points $i$ and $i+1$)}
      \label{fig:calib-jig}
   \end{figure}

Depending on the application and the required accuracy, different researchers have designed and studied FBG-based shape sensors with various structures, different FBG placement and number of active areas \cite{park2010real}-\cite{wang2016shape}. In \cite{sefati2016fbg}, we proposed a design for CDM shape sensing by embedding a fiber array with three FBG active areas and two nitinol wires as substrates into a three-lumen Polycarbonate tube. In this paper, we continue the work by manufacturing and embedding two shape sensors in two channels on opposite sides of the CDM (Fig. \ref{fig:CDM}) and reconstructing the shape of the CDM to find its tip position. The shape sensor design is much more robust compared to the previously studied shape sensors, since the three-lumen Polycarbonate tube enables accurate placement of the FBGs and the nitinol wires, resulting in a more accurate sensor bias. The shape reconstruction method is as follows. When the CDM is bent, wavelength shift of active areas on the fiber are related to the strain changes in the shape sensors. Given the strain value, curvature is then calculated by:

\begin{figure}[t]
      \centering
      \includegraphics[scale=0.31]{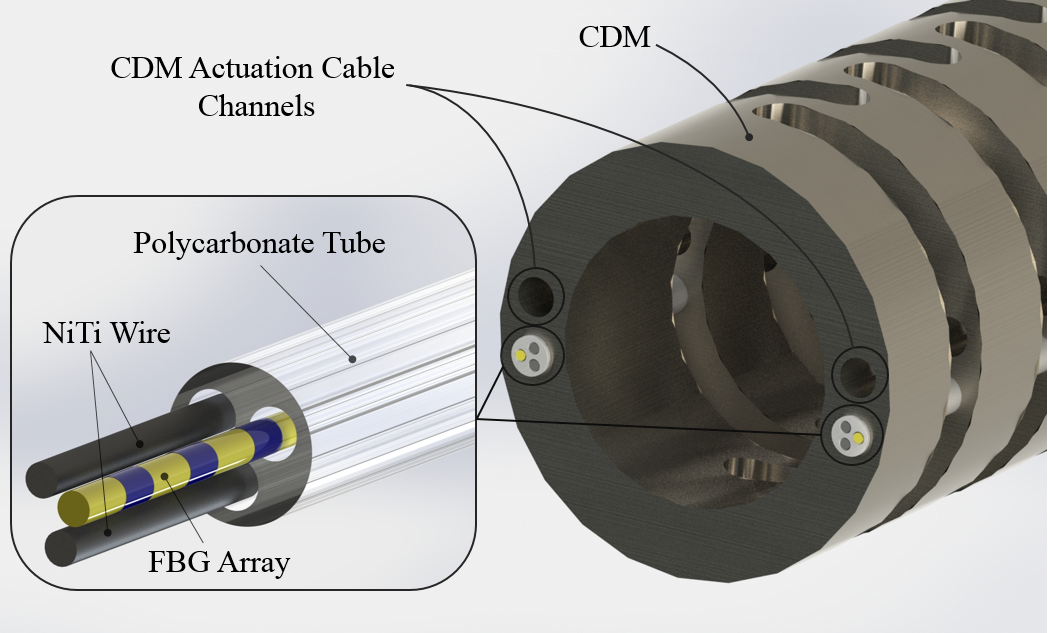}
      \caption{View of the distal end (tip) of the CDM, indicating the actuation cable channels, and the Sensor assembles embedded in the sensor channels.}
      \label{fig:CDM}
\end{figure}


\begin{equation} \label{eq:strain-wave}
	\frac{\Delta \lambda}{\lambda_B} = k_\epsilon \epsilon + k_T \Delta T
\end{equation}
where $\Delta T$, $\kappa$, $\epsilon$, $\Delta \lambda$, and $\lambda_B$ refer to the temperature variation, curvature, strain, wavelength shift and Bragg wavelength of the FBGs, respectively. $k_\epsilon$ and $k_T$ are the strain and temperature coefficients. Assuming the temperature variation is very small and negligible ($\Delta T \approx 0$), the changes in wavelength shift is due to the changes in mechanical strain, resulting from the bending stress. This mechanical strain is related to the curvature by:
\begin{equation} \label{eq:strain-curvature}
	\epsilon = \delta \kappa
\end{equation}
where $\kappa$ and $\delta$ refer to curvature and sensor bias, respectively. Combining (\ref{eq:strain-wave}) and (\ref{eq:strain-curvature}) and assuming $\Delta T \approx 0$, we obtain:

\begin{equation} \label{eq:wave-curvature}
	\kappa = \frac{\Delta \lambda}{\lambda_B k_\epsilon \delta}
\end{equation}
where $\lambda_B$, $k_\epsilon$ and $\delta$ are all constants. Therefore, we can assume a linear relationship between $\Delta \lambda$ (wavelength shift) and $\kappa$ (curvature) at each active area of the FBG shape sensors. $\lambda_B$ (Bragg wavelength) and $k_\epsilon$ (mechanical strain constants) are known from properties of the FBGs. On the other hand, $\delta$ (sensor bias) depends on how accurate and at what distance from the center of the Polycarbonate tube, the FBGs are glued. In theory, this value is equal to the distance of the center of the lumens from the center of the Polycarbonate tube. However, in practice, due to small clearance between the fibers and the lumen, this distance cannot be measured accurately. For this reason, we perform a calibration procedure to find the mapping between the FBG wavelength shifts and curvature values:

\begin{equation} \label{eq:calibration}
	\kappa = f(\Delta \lambda)
\end{equation}
where $f: \mathbb{R}^3 \rightarrow \mathbb{R}^3$ is a linear function relating the wavelength shift at the three active areas of each shape sensor to curvature values at these three points. To find $f$, we have $3$D printed calibration jigs with known constant curvatures for the CDM (Fig. \ref{fig:calib-jig}-a). The curvature in these jigs vary from straight to a radius of curvature of $20$ mm. The FBG wavelength data is streamed by a dynamic optical sensing interrogator (Micron Optics sm 130) at frequency of $200$ Hz. The CDM is placed into the jigs of Fig. \ref{fig:calib-jig} and the wavelength data is collected by the interrogator. A linear function is fit to the wavelength data and the known jig curvatures to find the calibration mapping $f$. Then any real-time wavelength data is passed to the function $f$ to find the curvature at the location of each of the active areas. 
  
   \begin{figure}[bt]
      \centering
      \includegraphics[scale=0.31]{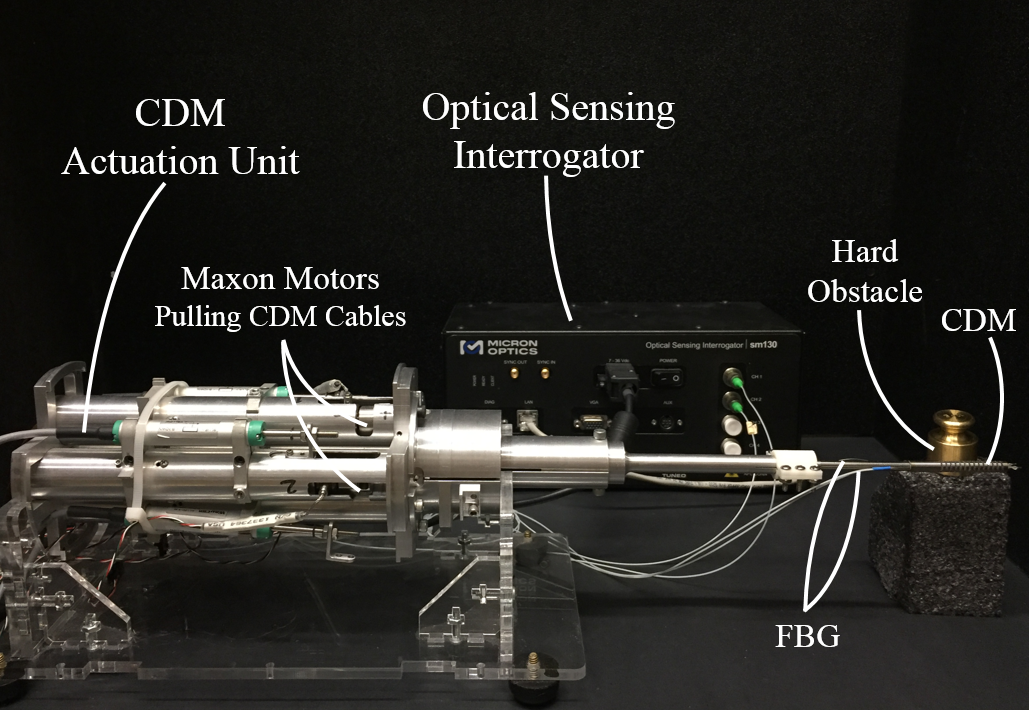}
      \caption{The experimental setup.}
      \label{fig:setup}
   \end{figure}
   
Using the calibration function $f$, the curvature is found at three discrete points along the length of the CDM on each shape sensor. We assume a linear relation between the arc-length and curvature and find the discretized curvature for a specific number of points (e.g. $10$) on each segment between two FBGs. The angle of curvature between each two consecutive points is related to the curvature and arc-length by:

\begin{equation} \label{eq:curvature-arc}
\begin{split}
	\kappa &= a s + b \\
    \Delta\theta_i = &\kappa_i \Delta s = \frac{\Delta s}{\rho_i} 
\end{split}
\end{equation}
where $s$ is the arc-length, $a$ and $b$ are constants, $\rho$ is the radius of curvature and $\theta$ is the angle of curvature. The $2$D position ($y$ and $z$) along the FBG shape sensor can then be found by (Fig. \ref{fig:calib-jig}-b):

\begin{equation} \label{eq:yz}
\begin{split}
	y_{i+1} &= y_i + \rho_i \sin(\Delta \theta_i) \\
	z_{i+1} &= z_i + \rho_i (1 - \cos(\Delta \theta_i))
\end{split}
\end{equation}

Using (\ref{eq:curvature-arc}) and (\ref{eq:yz}) for each shape sensor, we can obtain the position of shape sensor and the tip position of the CDM by averaging the position values for the shape sensors which we refer to as $x \in \mathbb{R}^2$.
   
\subsection{Optimization-based Control Framework}

The manipulator Jacobian ($J$) is a configuration-dependent function that maps the actuation input velocities ($\dot{l}$) to the end-effector velocities ($\dot{x}$):
\begin{equation} \label{eq:Jacobian}
\dot{x} = J \dot{l}
\end{equation}

Using (\ref{eq:Jacobian}), for each time step ($k$), the changes in the end-effector position for an infinitesimal time period can be estimated based on the changes in the actuation input: 

\begin{equation}\label{eq:deltaJacobian}
J_k \approx \frac{\Delta x_k}{\Delta l_k}.
\end{equation}
where $\Delta l_k$ and $\Delta x_k$ are the changes in the actuation inputs and the end-effector displacements, respectively. Considering (\ref{eq:deltaJacobian}), one reasonable strategy for controlling the CDM to a target point is to find the incremental actuation input that moves the CDM end-effector to the target point. Further, as we discussed, real world problems have various constraints, e.g. the CDM actuation input might be subject to some motion constraints. We, therefore, formulate the following optimization to address this problem: 

\begin{equation} \label{eq:input-update}
\begin{aligned}
& \underset{\Delta l_k}{\text{minimize}}
& & \|\Delta x_{des} - J_k \: \Delta l_k\| \\
& \text{subject to}
& & \: \: \: \: \: \: A \: \Delta l_k \leq b
\end{aligned}
\end{equation}
where $\Delta x_{des}$ is the end-effector required tip position displacement to reach the desired goal point, and $A \in \mathbb{R}^{m\times n}$ and $b \in \mathbb{R}^{m \times 1}$ denote $m$ inequality constraints  on $\Delta l_k$ ($\mathbb{R}^n$), at the time step (k). These inequality constraints define the feasible region for the optimization variable $\Delta l_k$, which is crucial considering the possible limitations in the physical application. 

In each time step $k$, given the actuation control input calculated from (\ref{eq:input-update}), the CDM is moved. Then the correct Jacobian that would have made the movement of $\Delta x_k$ from an actuation control input of $\Delta l_k$ is calculated using (\ref{eq:Jacobian-update}). To make a smoother transition from the current Jacobian at the time step (k) to the next step, we minimize the Frobenius matrix norm of $\Delta J$:
\begin{equation} \label{eq:Jacobian-update}
\begin{aligned}
& \underset{\Delta J_{k+1}} {\text{minimize}}
& & \|\Delta J \| \\
& \text{subject to}
& & \Delta x_k = J_{k+1} \: \Delta l_k \\
&&& J_{k+1} = J_k + \Delta J
\end{aligned}
\end{equation}
where $\Delta l_k$ is the calculated actuation control input from (\ref{eq:input-update}) and $\Delta x_k$ is the end-effector tip position displacement read from the FBG sensors between the last Jacobian estimate at time step $k$ and the next time step. A threshold ($\epsilon$) can be specified as the termination factor of the algorithm. The size of this threshold can be defined experimentally and based on the noise of the sensor feedback. In other words, the control input and the Jacobian will be updated as long as $\Delta x_{des}$ is greater than this threshold value. It should be noted that a major advantage of estimating the jacobian on the fly using Eq. \ref{eq:Jacobian-update} is that regardless of the type of environment surrounding the CDM (e.g. free space or with obstacle interactions), the CDM actuation inputs for next step is computed accordingly by incorporating information about the CDM behavior in that environment. Algorithm (\ref{control-algorithm}) summarizes the proposed control strategy.

   \begin{algorithm}[tb] \label{control-algorithm}
    \SetKwInOut{Input}{Input}
    \SetKwInOut{Output}{Output}    
    \Input{Jacobian at time step k, $J_k$ \\ 
    Target tip position, $x_{target}$ \\
    Threshold, $\epsilon$} 
    \Output{Actuation input $\Delta l_k$ \\ 
    Jacobian at the next step}
    Query $x_{current}$ from FBG sensors \\
    Compute $\Delta x_{desired} = x_{target} - x_{current} $ \\
    

	\While{$\Delta x_{desired} \geq \epsilon$}{
	$ \begin{aligned}
	& \Delta l_k \leftarrow {\text{minimize}}
	& & \|\Delta x_{des} - J_k \: \Delta l_k\| \\
	&  \: \: \: \: \: \: \: \: \: \: \: \: \: \: \: \: \text{subject to}
	& & \: \: \: \: \: \: A \: \Delta l_k \leq b
	\end{aligned}$
    
    $\begin{aligned}
	& \Delta J_{k+1} \leftarrow {\text{minimize}}
	& &  \: \: \: \: \:  \|\Delta J \| \\
	& \: \: \: \: \: \: \: \: \: \: \: \: \: \: \: \:  \: \: \: \: \text{subject to}
	& & \Delta x_k = J_{k+1} \: \Delta l_k \\
	&&& J_{k+1} = J_k + \Delta J
	\end{aligned}$
    }
    \caption{Optimization-based Control Algorithm}
\end{algorithm}

\section{Experiments}

\subsection{Experimental Setup} \label{setup}

The proposed planar CDM for the treatment of osteolysis, consists of two actuation cables on the side channels. The CDM cables are actuated with two DC motors (RE10, Maxon Motor Inc. Switzerland) with spindle drives (GP 10 A, Maxon Motor, Inc. Switzerland). A commercial controller is used to power and connect individual Maxon controllers (EPOS 2, Maxon Motor Inc. Switzerland) on a CAN bus. Using libraries provided by Maxon, a custom C++ interface communicates over a single USB cable and performs position control of the motors. Fig. \ref{fig:setup} shows the experimental setup. 

FBG data is streamed by a dynamic optical sensing interrogator (Micron Optics sm 130) at frequency of $200$ Hz. Two fibers, each with three active areas, are connected to separate channels of the interrogator. These data are used to reconstruct the shape of the CDM and find the tip position ($x$) using the method described in section \ref{Feedback}.
   
\subsection{Constraints}
Considering (\ref{eq:input-update}), we need to define $A$ and $b$ based on the constraints imposed by the application. The CDM is capable of bending to large curvatures up to $166.7 m^{-1}$ \cite{murphy2014design}. Therefore, there is a maximum allowable cable actuation that prevents the CDM from breaking. In addition, to avoid large actuation inputs in each step of (\ref{eq:input-update}), we limit the displacement of the cables. Considering the constraints, we can define matrix $A$ and vector $b$ as:

\[
A
=
\begin{bmatrix}
    1 &  0 & -1 & 0 & 1 & -1 & 1 & 0\\
    0 &  1 & 0 & -1 & 1 & -1 & 0 & 1\\
\end{bmatrix}^T \]
\[
b
=
\begin{bmatrix}
    1 & 1 & 1 & 1 & 0.1 & 0.1 & b_1 &b_2\\
\end{bmatrix}^T
\]
\begin{equation*}
b_1 = b_2 = l_c - l_{max}
\end{equation*}
where $l_{max}$ and $l_c$ are the maximum allowable cable length and the cable length at the current iteration, respectively. The first four rows in $A$ and $b$ ensure that the change in string length in each iteration is less than $1$ mm. Rows five and six will make the sum of changes in both string lengths to be less than $0.1$ mm (i.e., ensure both cables have approximately the same amount of movement to avoid excessive counter-tensioning). These constraints account for the pull-pull behavior of the two cables which ensures that the cables are not interfering with each other. The last two rows guarantee that the achieved length of the cables at the next iteration of the algorithm does not exceed the maximum allowable length. $l_{max}$ is chosen $7$ mm based on limitations of the experimental setup.   

      \begin{figure}[bt]
      \centering
      \includegraphics[scale=0.36]{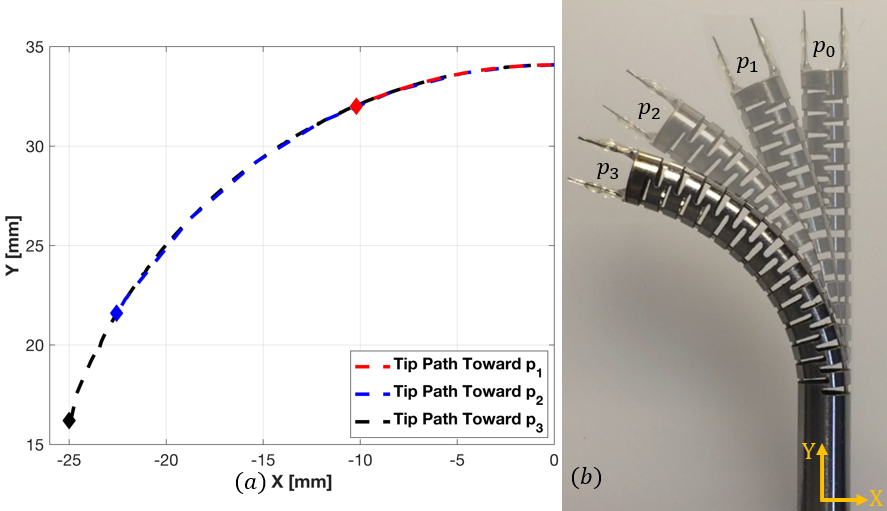}
      \caption{a) The CDM tip traversed path for three target points in free bending environment, b) the CDM instantaneous configuration when the tip has reached the target points.}
      \label{fig:free-bends}
   \end{figure}

\subsection{Experiments}

All experiments are run on a 64-bit Windows 7 computer with an Intel 2.3 GHz core i7 processor with 8GB RAM. The FBG data streaming occurs at frequency of $200$ Hz and the control loop runs at frequency of $100$ Hz. The two-norm convex objective functions and the constraints of the optimization problems were transformed to a simple quadratic program with linear equality constraints. For the control input optimization, we rewrote the vector two-norms as multiplication of the vector by its transpose. For the Jacobian update problem, we stacked the elements of the $\Delta J$ matrix in a vector and minimized the two-norm of this vector. We used the C++ QuadProg\footnote{http://www.diegm.uniud.it/digaspero/} library---a quadratic optimization solver---to solve the optimization problems of (\ref{eq:input-update}) and (\ref{eq:Jacobian-update}).

We chose $0.05$ mm/s as the cable actuating velocity for all of the experiments. This velocity agrees with the results of \cite{alambeigi2016design}, where we investigated the optimal velocity for debriding hard and soft osteolytic lesions using appropriate debriding tools. In addition, a termination threshold value of $\epsilon = 0.05$ mm was chosen in all experiments. We also considered the Frobenius norm to evaluate the changes in the adapted Jacobian in each time step.

During the treatment of osteolytic lesions, the CDM may interact with different environments. We, therefore, considered three types of experiments to simulate these situations: 1) moving the CDM in a free environment, 2) interaction of the CDM with soft obstacles mimicking soft tissues, and 3) interaction of the CDM with hard obstacles (e.g., bone or sclerotic tissue).

\subsubsection{Free Environment Bending} \label{freeBend}
We first tested the behavior of the algorithm for the CDM control in an environment without any obstacles. In this experiment, the controller's ability to move the CDM tip to three different target points was evaluated (Fig. \ref{fig:free-bends}). We chose target points $p_1 = [-10, 32.1]$ mm, $p_2 = [-22.6, 21.4]$ mm, and $p_3 = [-25, 16.2]$ mm, corresponding to different levels of bending. In addition, to ensure repeatability, the CDM performed the following moving sequence: 1) from straight configuration ($p_0 = [0, 34]$) to the target position $p_1$; 2) from position $p_1$ to the straight configuration $p_0$; 3) from straight configuration $p_0$ to target position $p_1$; and 4) from position $p_1$ to the straight configuration $p_0$.
\subsubsection{Soft Obstacle Environment}\label{soft}
\begin{figure}[t]
      \centering
      \includegraphics[scale=0.36]{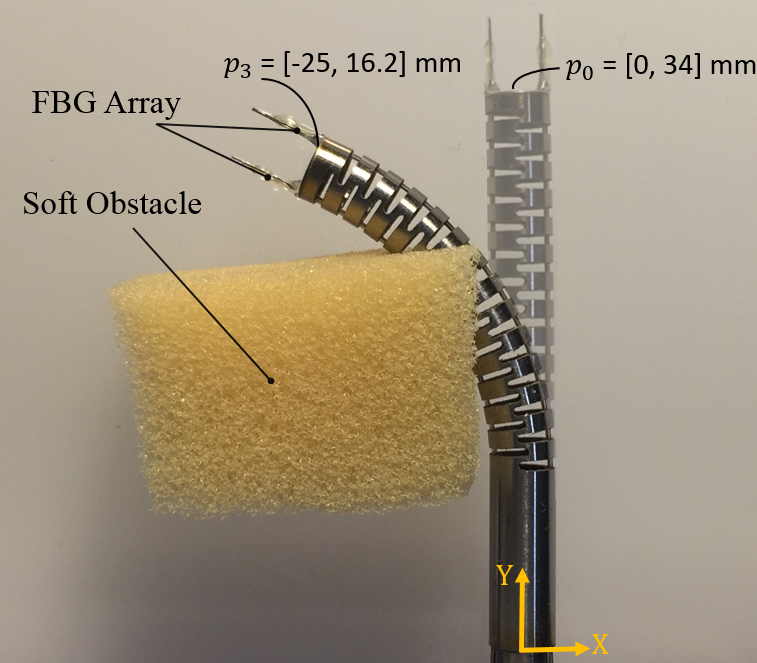}
      \caption{ Interaction of the CDM with soft obstacle.}
      \label{fig:soft}
   \end{figure}

To study the behavior of the controller for the interaction of the CDM with soft environments, we arbitrarily placed a soft object in the bending path of the CDM (Fig. \ref{fig:soft}). To compare the behavior of the CDM with the free bending case, we chose $p_3$ as the target point.

\subsubsection{Hard Obstacle Environment}\label{hard}
Similar to the soft obstacle experiment, we arbitrarily placed a rigid object in the bending path of the CDM. To compare the bending behavior of the CDM as well as the controller performance with the aforementioned cases (i.e free environment and soft obstacle), we chose $p_3$ as the target point.

\section{Results and Discussion}
Fig. \ref{fig:free-bends} shows the CDM configurations and the traversed tip paths as it is moving from the straight configuration ($p_0$) toward the three aforementioned target points ($p_1, p_2, and \, p_3$) in the free bending environment. Due to the variable-curvature behavior of the CDM \cite{murphy2014design}, its dynamics alters as the CDM is undergoing different levels of bending. As observed in Fig. \ref{fig:free-bends}, the proposed control algorithm can address this change of dynamics by guiding the CDM toward the desired targets. Further, similar traversed paths in different experiments confirm repeatability of the CDM bending behavior and show negligible hysteresis. 

   \begin{figure}[t]
      \centering
      \includegraphics[scale=0.58]{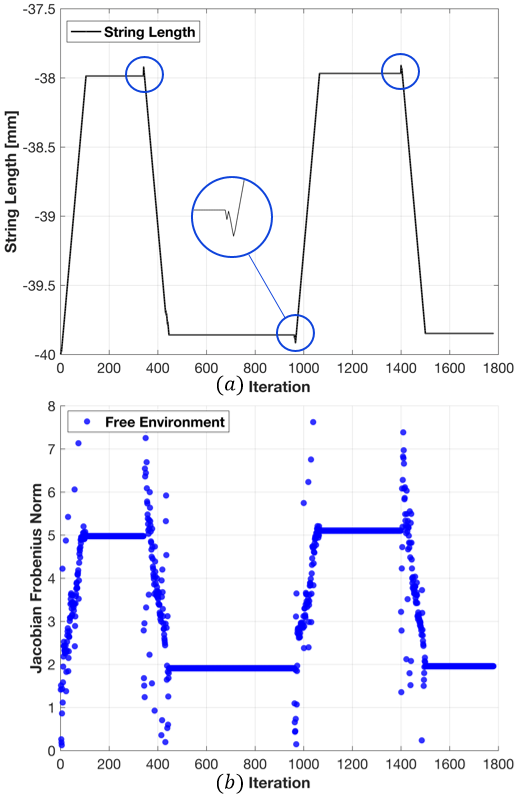}
      \caption{ Repeatability experiment in free environment a) changes in string length over time b) the estimated Jacobian over time.}
      \label{fig:repetition}
   \end{figure}
 
   \begin{figure}[t]
      \centering
      \includegraphics[scale=0.535]{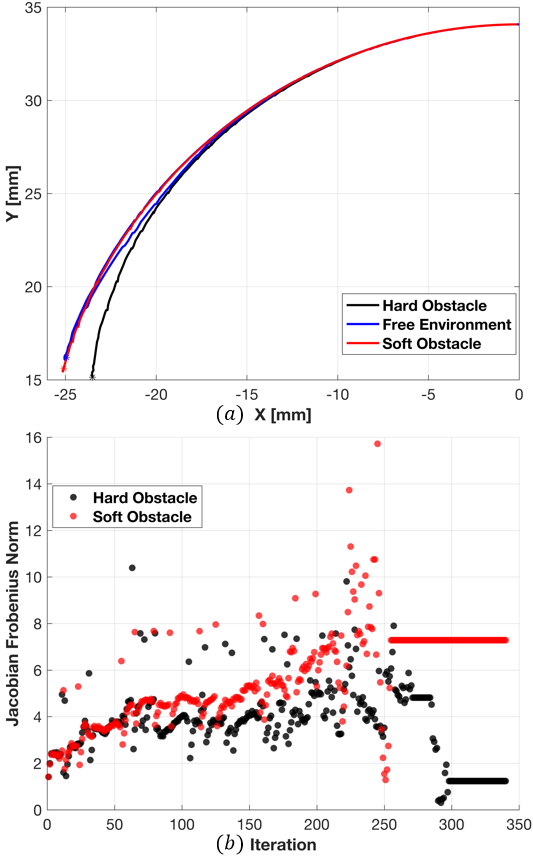}
      \caption{a) The CDM tip transversed path for target point $p_3 = [-25, 16.2]$ mm in free bending, soft obstacle, and hard obstacle experiments b) the estimated Jacobian comparison in soft and hard obstacle experiments.}
      \label{fig:hard-soft-free}
   \end{figure}
   
Fig. \ref{fig:repetition} indicates the repeatability of the proposed method in terms of the control input and norm of the adapted Jacobian matrix. As shown, similar patterns are observed both in control input and norm of the Jacobian for the repeated experiments. In both figures, the rising parts of the plots demonstrate the bending motion of the CDM toward the target point $p_1$, while the falling parts indicate the movement toward the target point $p_0$. Investigation of the control input demonstrates small peaks at the beginning of each rising and falling part (blue circles in Fig. \ref{fig:repetition}-a). This behavior illustrates the learning step of the proposed control method, where the controller tends to obtain the correct movement direction toward the target point. Further, similar behavior is observed in the norm of the calculated Jacobian (Fig. \ref{fig:repetition}-b), supporting the learning step behavior. In addition, values of the Jacobian norm increased as the CDM moved toward $p_1$ and decreased when moved toward $p_0$. This means that in a highly bent configuration of the CDM, a specific control input (i.e. change of the cable lengths) results in a larger displacement of the CDM tip as compared to the less bent configurations.

Fig. \ref{fig:hard-soft-free}-a compares the CDM traversed tip paths---reported by the FBGs---during movement toward the target point $p_3$ in the soft obstacle, the hard obstacle and the free bending experiments. As shown, regardless of the external contact forces exerted by the soft obstacle, the CDM traversed approximately similar paths to the free bending experiment. This indicates that the proposed method can compute effective control inputs to overcome the unknown disturbance imposed by the contact force. Further, in the hard obstacle experiment, we observe higher deviation from the free bending traversed path (as expected), since the CDM cannot penetrate the hard obstacle, as opposed to the soft obstacle case (Fig. \ref{fig:hard} and Fig. \ref{fig:soft}). The higher error in the hard obstacle experiment between the desired target point and the final achieved tip position is because the CDM cannot physically reach the target point.

\addtolength{\textheight}{-2cm}  

Fig. \ref{fig:hard-soft-free}-b demonstrates the norm of the adapted Jacobians over time for both of the soft and hard obstacle experiments. These plots show similar behavior before iteration $60$, where there is no contact between the CDM and obstacles. After collision, the norm of the Jacobian changes; in the soft obstacle case the norm of the Jacobian is larger compared the other case. A higher reduction in the Jacobian norm for the hard obstacle interaction demonstrates a higher contact force compared to the soft obstacle interaction. 
Furthermore, the last movements in the hard obstacle experiment (i.e. iterations $\geq 280$) demonstrate a reduction in the Jacobian norms. Physically, these iterations correspond to the situation when the CDM is wrapping around the cylindrical obstacle and cannot proceed toward the desired target (Fig. \ref{fig:hard}). In this case, any specific input command (i.e. string length) results in a small CDM tip displacement.

It should be noted that Wilkening et. al. \cite{wilkening} have previously used a constant jacobian matrix for controlling the CDM similar to the one used in this study using a PD controller. The jacobian was found by pulling the actuation cables during free bending motion and observing tip behavior. However, a constant experimental jacobian would limit the control strategy to only be valid during free space bending of the CDM (and not with unknown obstacle interactions), which is not the case for the application of osteolysis.

\section{CONCLUSION}

In this paper, we proposed an optimization-based method to control a continuum manipulator, designed for less-invasive treatment of osteolysis, using real-time feedback from a Fiber Bragg Grating sensing unit and incorporating the mechanical constraints on the CDM imposed by the application. We used the provided FBG data to reconstruct the shape and estimate the tip of the CDM. To evaluate the performance of the proposed method, we designed experiments to mimic the real scenario of the CDM interactions with the soft and hard lesions as well as the free bending motion. The obtained results confirms successful control of the CDM tip in unknown environments (soft and hard obstacles). Further, the results indicated repeatability and robustness of the proposed method in controlling the CDM tip, though further experiments are necessary. In addition, we showed that with the proposed method, we are able to detect the CDM collisions with objects. Future works will focus on using the proposed algorithm to steer the CDM through a constraint environment. Moreover, the combined control of an integrated robotic system including the CDM and a robotic arm will be investigated.






\end{document}